\documentclass[10pt,twocolumn,letterpaper]{article}

\usepackage{iccv}
\usepackage{times}
\usepackage{epsfig}
\usepackage{graphicx}
\usepackage{amsmath}
\usepackage{amssymb}

\usepackage{xcolor}
\definecolor{mygray}{gray}{0.4}
\usepackage{mathtools}
\usepackage{acronym}
\usepackage{caption}
\usepackage{array}
\usepackage{tabularx}
\usepackage{subcaption}
\usepackage{booktabs}
\usepackage{multicol}
\usepackage{multirow}
\usepackage[norelsize, linesnumbered, ruled, lined, boxed]{algorithm2e}
\usepackage{url}
\usepackage{color}
\usepackage{xspace}
\usepackage{overpic}

\newcommand{\genBG}{\ensuremath{\mathcal{G}_\text{bg}}\xspace}

\newcommand{\gen}{\ensuremath{\mathcal{G}}\xspace}
\newcommand{\disc}{\ensuremath{\mathcal{D}}\xspace}
\newcommand{\alphanet}{\ensuremath{\mathcal{A}}\xspace}

\usepackage[pagebackref=true,breaklinks=true,letterpaper=true,colorlinks,bookmarks=false]{hyperref}
\iccvfinalcopy % *** Uncomment this line for the final submission

\ificcvfinal\fi

\begin{document}

%%%%%%%%% TITLE
\title{ Labels4Free: Unsupervised Segmentation using StyleGAN}

\author{Rameen Abdal\textsuperscript{1} \quad Peihao Zhu\textsuperscript{1} \quad Niloy Mitra\textsuperscript{2} \quad Peter Wonka\textsuperscript{1} \\
\\
\textsuperscript{1}KAUST \quad \textsuperscript{2}UCL, Adobe Research
}

\maketitle
\ificcvfinal\thispagestyle{empty}\fi

%%%%%%%%% ABSTRACT
\begin{abstract}
  We propose an unsupervised segmentation framework for StyleGAN generated objects. We build on two main observations. First, the features generated by StyleGAN hold valuable information that can be utilized towards training segmentation networks. Second, the foreground and background can often be treated to be largely independent and be composited in different ways. For our solution, we propose to augment the StyleGAN2 generator architecture with a segmentation branch and to split the generator into a foreground and background network. This enables us to generate soft segmentation masks for the foreground object in an unsupervised fashion. On multiple object classes, we report comparable results against state-of-the-art supervised segmentation networks, while against the best unsupervised segmentation approach we demonstrate a clear improvement, both in qualitative and quantitative metrics. \\ Project Page :  \href{https:/rameenabdal.github.io/Labels4Free}{https:/rameenabdal.github.io/Labels4Free} 
\end{abstract}

\section{Introduction}
Given the high quality and photo-realistic results of current generative adversarial networks (GANs), we are witnessing their widespread adaptation for many applications. Examples include various image and video editing tasks, image inpainting~\cite{demir2018patch,yu2018generative,yu2018freeform,webster2019detecting}, local image editing~\cite{wu2020stylespace, abdal2020image2stylegan++}, low bit-rate video conferencing~\cite{wang2020one}, image super resolution~\cite{menon2020pulse,daras2021intermediate}, image colorization~\cite{afifi2020histogan,luo2020time}, and extracting 3D models~\cite{pan2020gan2shape}.

\begin{figure}[t!]
    \centering
    \includegraphics[width=\linewidth]{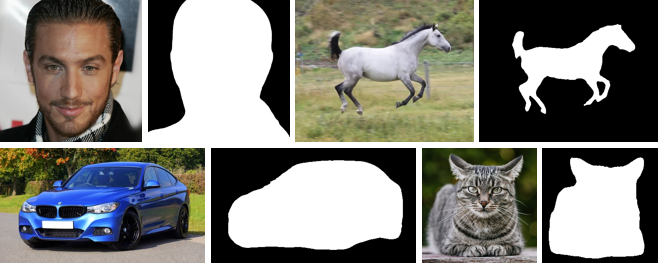}
    \caption{We propose an unsupervised segmentation framework that enables foreground/background separation for raw input images. At the core of our framework is an unsupervised network,  which segments class-specific StyleGAN images, and is used to generate segmentation masks for training supervised segmentation networks.}
    \label{fig:teaser}
\end{figure}

While originally conjectured that GANs are merely great at memorizing the training data, recent work in GAN-based image editing~\cite{harkonen2020ganspace,abdal2020styleflow,shen2020interfacegan, wu2020stylespace}  demonstrates that GANs learn non-trivial semantic information about a class of objects, e.g., faces or cars. For example, GANs are able to learn the concept of pose and they can show the same, or at least a very similar looking, object with different orientations. Even though the background changes in subtle ways during this editing operations, in this paper we explore to what extent the underlying  generator network actually learns the distinction between foreground and background, and then to encourage it to disentangle foreground and background, without explicit mask-level supervision. As an important byproduct, we can extract information from an unsupervised GAN that is useful for general object segmentation. For example, we are able to create a large synthetic dataset for training a state-of-the-art segmentation network and then segment arbitrary face (or horse, car, cat) images into foreground and background \textit{without} requiring any manually assigned labels (see Fig.~\ref{fig:teaser}).

Our implementation is based on StyleGAN~\cite{Karras2019stylegan2, karras2019style}, generally considered the state-of-the-art for GANs trained on individual object classes. Our solution is built on two ideas. First, based on our analysis of GAN-based image editing, the features generated by StyleGAN hold a lot of information useful for segmentation, and can be used towards corresponding mask synthesis. 
Second, the foreground and background should be largely independent and be composited in different ways. The exact coupling between foreground and background is highly non-trivial however and there are multiple ways of decoupling foreground and background that we analyze in our work.

For our solution, we propose to augment the StyleGAN generator architecture with a segmentation branch and to split the generator into a foreground and background network. This enables us to generate soft segmentation masks for the foreground object. In order to facilitate easier training, we propose a training strategy that starts from a fully trained network that only has a single generator and utilizes it towards unsupervised segmentation mask generation. 

To summarize, our main contributions are:
\vspace*{-.5em}
\begin{itemize}
  \setlength\itemsep{0em}
    \item A novel architecture modification, a loss function, and a training strategy to split StyleGAN into a foreground and background network.
    \item Generating synthetic datasets for segmentation. Our framework can be used to create a complete dataset of labeled GAN generated images of high quality in an unsupervised manner. This dataset can then be used to train other state-of-the-art segmentation networks to yield compelling results.
\end{itemize}

\if0
Despite producing high quality samples, the interpretability and controllability of neural networks and particularly GANs, is still an active area of research.
In many recent papers, researchers discovered that GANs are able to learn interesting and useful semantic concepts in an unsupervised manner. In many cases it has been more successful to extract semantic concepts from a pre-trained generator, rather than modifying the architecture. 

In this paper,
we tackle a general interpretability problem of GANs. We study the composition of the layers of StyleGAN and study their contribution to the final output. In particular, we investigate the foreground/background separation capabilities of StyleGAN without mask level supervision. This leads to insights into the compatibility properties of GANs and potential to be used as an unsupervised 3D scene compositor. The analysis opens many potential applications in image editing and image restoration.
\fi

\section{Related Work}

\paragraph{High Quality GANs.}
From the seminal works of Goodfellow et al.~\cite{goodfellow2014generative} and Radford et al.~\cite{radford2015unsupervised}, subsequent GAN research contributed to big improvements in the visual quality of the generated results. The state-of-the-art networks like ProGAN~\cite{karras2018progressive}, BigGAN~\cite{brock2018large}, StyleGAN~\cite{karras2019style}, StyleGAN2~\cite{Karras2019stylegan2} and StyleGAN2-ada~\cite{Karras2020ada} demonstrate superior performance in terms of diversity and quality of the generated samples. While StyleGAN series by Karras et al.~\cite{Karras2020ada} has demonstrated high quality and photo-realistic results on human faces using the high quality FFHQ~\cite{karras2019style} dataset, BigGAN can produce high quality samples using complex datasets like ImageNet. In our work we build on StyleGAN2 which is the current state of the art for many smaller data-sets, including faces.

\paragraph{GAN Interpretability and Image Editing.}
GAN interpretability has been an important aspect of the GAN research since the beginning. Some recent works in this domain~\cite{bau2018gan,bau2019seeing, fruhstuck2019tilegan, harkonen2020ganspace, zhu2020improved, abdal2020styleflow, shen2020interfacegan, abdal2020image2stylegan++, tewari2020stylerig} study the structure of the activation and latent space. For instance GANspace simplifies the latent space of StyleGAN ($W$ Space) to be linear and extracts meaningful directions using PCA. StyleRig~\cite{tewari2020stylerig} mapped StyleGAN latent space to a riggable face model. StyleFlow~\cite{abdal2020styleflow} studied the non-linear nature of the StyleGAN latent space using normalizing flows and is able produce high-quality sequential edits on the generated and real images.  On the other hand, the layer activation based methods~\cite{fruhstuck2019tilegan, abdal2020image2stylegan++,bau2018gan,  wu2020stylespace} try to understand the properties of GANs using an analysis of the activation space. A recent work StyleSpace~\cite{wu2020stylespace} studies the style parameters of the channels to determine various properties and editing abilities of StyleGAN.

Another interesting approach for interpretability of GANs is image embedding. Abdal et. al.~\cite{abdal2019image2stylegan} demonstrated high quality embeddings into the extended $W$ space called $W^{+}$ space for real image editing. Subsequent works~\cite{zhu2020domain,richardson2020encoding,tewari2020pie} try to improve upon the embedding quality, e.g. by suggesting new regularizers for the optimization. The embedding methods combined with image editing capabilities of StyleGAN has lead to many applications and even commercial software such as Adobe Photoshop's Neural Filters~\cite{Adobe}. A known problem in the domain of image editing is the disentanglement of the features during an editing operation. While state-of-the-art editing frameworks~\cite{abdal2020styleflow, tewari2020stylerig, harkonen2020ganspace} achieve high quality fine grained edits using supervised and unsupervised approaches, background/ foreground aware embeddings and edits still pose a challenge. Often, the background of the scene changes a lot during the embedding itself or during sequential edits performed on the images. Later in Sec.~\ref{Sec:Res}, we show how our unsupervised segmentation framework leads to improved image editing.

\paragraph{Unsupervised Object Segmentation.}
Among the few attempts in this domain are works by Xu et al.~\cite{ji2019invariant} and Yassine et al.~\cite{ouali2020autoregressive} which learn a clustering function in an unsupervised setting. A more recent work, Van et al.~\cite{van2021unsupervised} adopts a predetermined prior in a contrastive optimization
objective to learn pixel embeddings. In the GAN domain, PSeg~\cite{bielski2019emergence} is currently the only approach to segmenting foreground and background by reformulating and retraining the GAN using relative jittering between the composite and background generators at the cost of quality of samples.

%%%%%%%%% BODY TEXT
\section{Method}
\label{sec:meth}

Our goal is to segment StyleGAN generated images (i.e., a generative model that allows sampling from a normal distribution $z \sim \mathcal{N}(0,\mathbb{I})$ to produce class-specific  images $\mathcal{G}(z)$) into foreground and background layers. We achieve this \textit{without} access to ground truth masks. Our self-supervised algorithm relies on two observations: first, for many object classes, foreground and background layers can be swapped and recomposited to still produce good quality images;  and second, the interior layers of a trained StyleGAN already contain valuable information that can be used to extract such foreground/background segmentation.

\begin{figure}[t!]
    \centering
    \begin{overpic}[width=\columnwidth]{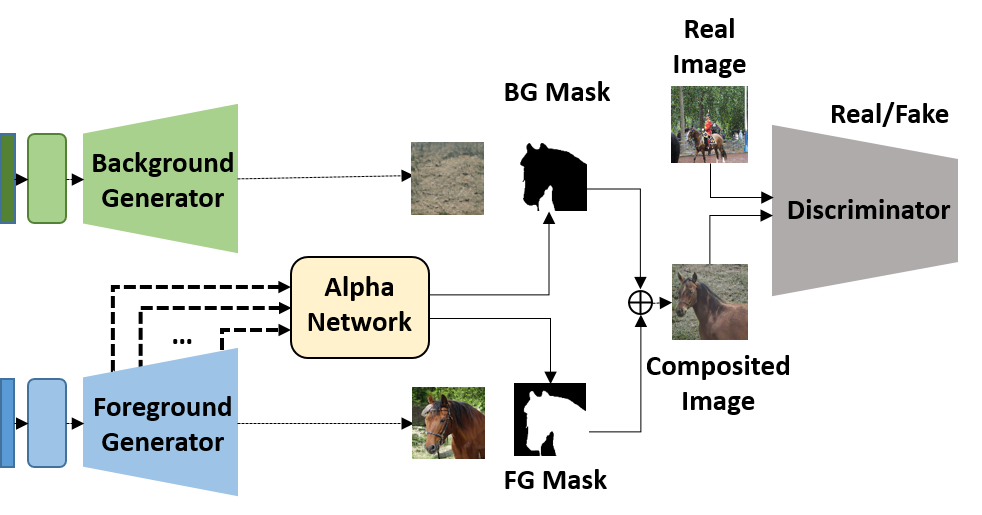}
    \small
    \put(12,0){$\gen$}
    \put(12,43){$\genBG$}
    \put(86,20){$\disc$}
    \put(34,12){$\alphanet$}
    \tiny
    \put(0,16){$\mathcal{N}(0,\mathbb{I})$}
    \put(0,41){$\mathcal{N}(0,\mathbb{I})$}
    \put(55,0){$M$}
    \put(51.5,39){$(1-M)$}
    \put(49.3,33.5){\small{$\odot$}}
    \put(49.3,8.5){\small{$\odot$}}
    \end{overpic}
    \caption{Our unsupervised segmentation network that makes use of pretrained generators \gen and \genBG to simultaneously train a segmentation network \alphanet (see Figure~\protect\ref{fig:frame2}) and a `weak' discriminator \disc, without requiring supervision for ground truth masks (see Sec~\ref{sec:meth}). } 
    \label{fig:frame}
\end{figure}

\begin{figure}[t!]
    \centering
    \includegraphics[width=.9\columnwidth]{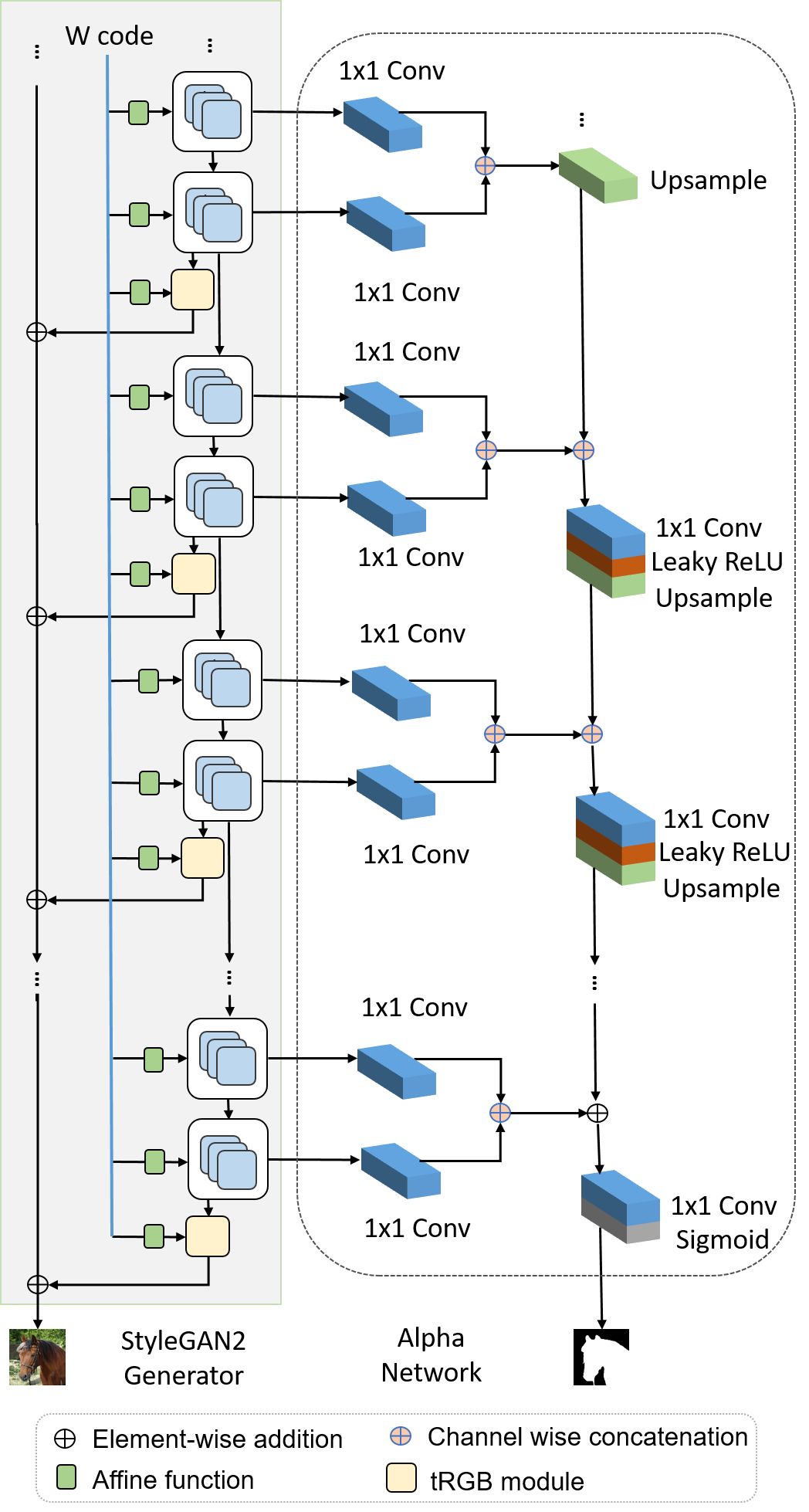}
    \caption{Architecture of the Alpha Network \alphanet that directly operates on pretrained features of StyleGAN to directly produce a binary segmentation mask $M$.}
    \label{fig:frame2}
\end{figure}

Our framework consists of two generators that share a single discriminator. The first generator  \gen is a pretrained StyleGAN generator whose channels are used by the AlphaGenerator Network \alphanet to eventually extract the foreground image as $\alphanet(z) \odot \gen(z)$ (see Sec.~\ref{Sec:alpha}).
The second generator \genBG, the background generator is responsible to generate background image samples $\genBG(z')$  from $z'\sim \mathcal{N}(0,\mathbb{I})$ (see Sec.~\ref{Sec: back}). Note that both are derived from pretrained generators. The final image, obtained by mixing background/foreground from different branches, is composited using standard $\alpha$-blending as $\alphanet(z) \odot \gen(z) + (1-\alphanet(z)) \odot \genBG(z')$, which a discriminator $\disc$ cannot tell apart from the real images. Note that only the network \alphanet is trained in this setup using the regular GAN loss. 

Fig.~\ref{fig:frame} shows an overview of this architecture.

We expand on the training dynamics of this network in Sec.~\ref{subsec:netDynamics}.

\subsection{Foreground Generator with Alpha Network}
\label{Sec:alpha}
Features of the pretrained generator \gen already contain sufficient information to segment image pixels into foreground and background. We extend the foreground generator with an Alpha Network~(\alphanet) to learn the alpha mask for the foreground (see Fig.~\ref{fig:frame2}). In this context, we overcome multiple challenges. 

First, the feature maps from lower layers of the generator need to be upsampled. To this effect, we introduce upsampling blocks in the alpha network. 

Second, the number of features per pixel is quite high (e.g., $9000+$ for StyleGAN2). We therefore use $1\times1$ convolutions for compression and feature selection. This encourages dropping the several features that do not contain segmentation information. 

Third, we optionally discard channels and complete layers using a semi-automatic analysis using the tRGB layers of the StyleGAN2. These layers are RGB channels conditioned on the output tensors with the W codes at each resolution contributing to the final image. We initialize the tRGB layers in different resolutions with noise $\mathcal{N}(0,\mathbb{I})$ at each pixel position and analyze the SSIM loss with respect to the original sample.
We average the results at each resolution and discard the ones with high SSIM scores. For example, for faces, the tRGB corresponding to the 4th layer of StyleGAN ($32\times32$ resolution) is retained as the average SSIM for the 3rd layer is about 25\% higher than those at the 4th layer. Hence we select all the latter tensors including the 4th layer tensors for the construction of $\alphanet$. Similarly, we identify the dominant layers to be 2nd for LSUN-Horse and LSUN-Cat and 3rd for the LSUN-Car (see Appendix B).

Fourth, the features per pixel have to be processed to output an alpha value. We therefore use several non-linear activation functions in the upsampling blocks and add a sigmoid function at the end to make the output in the range [0,1]. This yields a lightweight but simplified network to extract segmentation information. In particular, neighboring pixels do not interact with each other. 

\subsection{Background Generator}
\label{Sec: back}

The challenge for the background generator is the training dynamics. Specifically, when initializing with a pre-trained generator, the method fails because the background already includes the foreground object. When pre-training the background generator on a different dataset, our attempts failed because the discriminator can easily detect that the backgrounds are out of distribution (see Supplemental (Appendix A)). We therefore adopted an approach based on the conjecture that the foreground and background image are  already composited by StyleGAN.

We start out by seeking channels which are responsible for generating the background pixels in a StyleGAN image. Let $\gen$ be the StyleGAN generator and, $w$ and $n$ be the latent and noise variables, respectively. In order to bootstrap the network to identify the channels, we first identify and collect generic StyleGAN backgrounds from multiple sampled images by cropping. We notice that background having, for example, a white, black and blue tinge are common in-domain backgrounds. Then we find the gradient of the objective function: $\| (\gen(w,n) - x)\|_2^2$  with respect to the tensors in the StyleGAN2 layers at all resolutions, where $x$ is the upsampled background (crops). The above process allows identifying the layers which are most responsible for deleting an object (e.g., faces, cats, horses) from a composited StyleGAN2 image. 

\begin{figure*}
    \centering
    \includegraphics[width=0.90\linewidth]{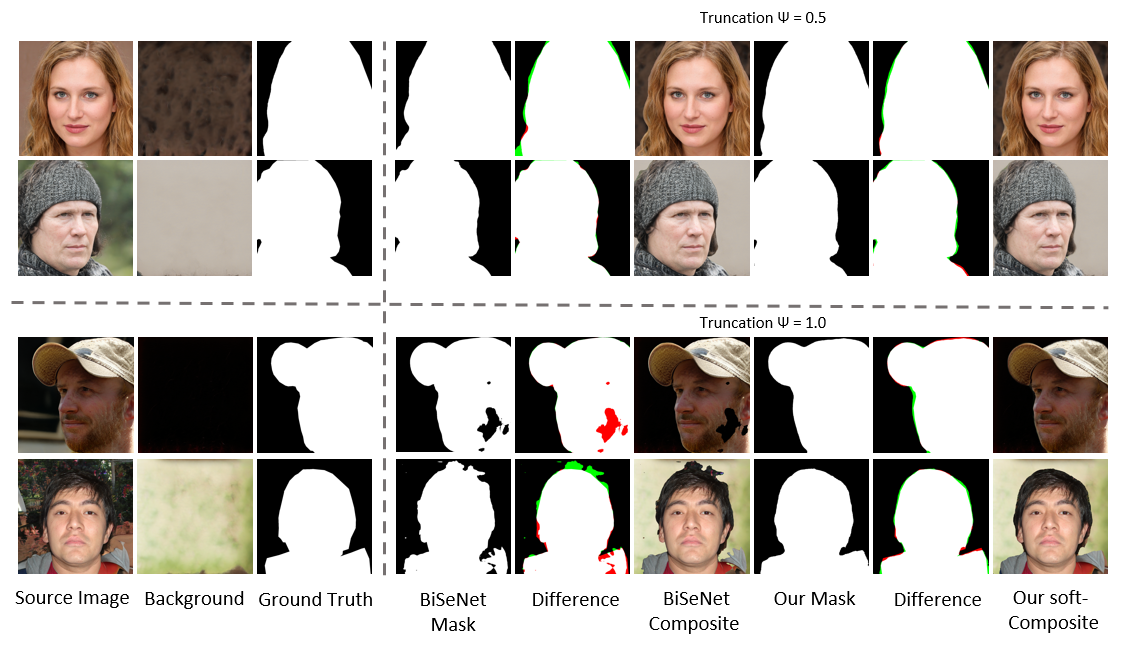}
    \caption{Qualitative results of our unsupervised framework on StyleGAN2 trained on FFHQ compared with BiSeNet trained on CelebA-HQ Masks. Note that the green and red areas are the `False Positives' and  `False Negatives' with respect to the foreground in ground truth.  We report our results on two truncation levels.
    } 
    \label{fig:fig1}
\end{figure*}

\begin{figure*}
    \centering
 
    \includegraphics[width=0.95\linewidth]{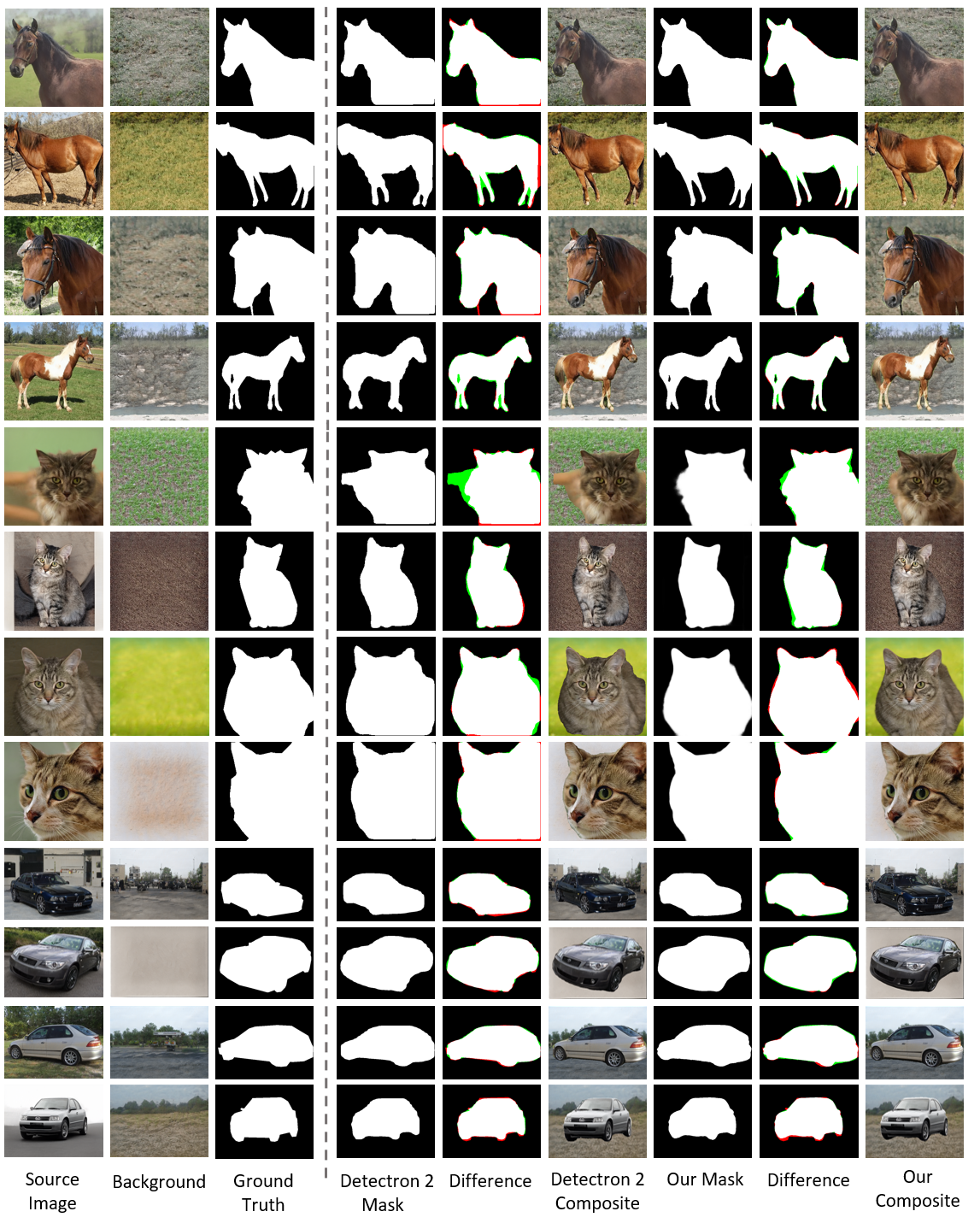}
    \caption{Qualitative results of our unsupervised framework on StyleGAN2 trained on LSUN-Horse, LSUN-Cat and LSUN-Car (LSUN-Object) compared with Detectron 2 trained on MS-COCO. Note that the green and red areas are the `False Positives' and  `False Negatives' with respect to the foreground in ground truth.} 
 
    \label{fig:fig_lsun_object}

\end{figure*}

In order to quantify the calculated gradient maps, we the calculate the sum of gradients norm over the channels of the respective layers. We found that the first layer (the constant layer and the first layer in which the W latent is injected excluding the tRGB layer~\cite{Karras2019stylegan2}) has the maximum value of the above measure. Hence, we hypothesise that switching off (i.e., zeroing out) the identified channels would exclude the object information from the tensor representation and produce an approximate background. Fig.~\ref{fig:fig1} and~\ref{fig:fig_lsun_object} show the sampled backgrounds following the steps described above. Note that we could  produce higher quality curated backgrounds by empirically setting some channels to be active based on a threshold of the above measure. However, we noticed that such backgrounds occasionally contained traces of foreground objects. Hence, in the training phase, we adopt a safe strategy and zero out all the channels of the selected layer. We refer to this trimmed background generator as $\genBG$.

\subsection{Training Dynamics}
\label{subsec:netDynamics}

Our unsupervised setup consists of the Alpha Network \alphanet, a pre-trained StyleGAN generator \gen for the foreground, a modified pre-trained StyleGAN generator $\genBG$ for the background, and a weak discriminator \disc. 

We freeze the generators \genBG and \gen during training and only train the Alpha Network \alphanet and the discriminator \disc using adversarial training. Note that we do not apply the path regularization to the discriminator \disc and style mixing during the training. Unlike others~\cite{bielski2019emergence}, our framework does not alter the generation quality of samples. The final composited image given to the Discriminator \disc is: 
\begin{equation}
    I(w,w', n, n') := M \odot \gen(w,n) + (1 - M) \odot \genBG(w',n'), 
\label{eq:composite}
\end{equation}
where $M$ is the mask predicted by the Alpha Network (i.e., $M = \alphanet(z)$). 
We use $W$ space for the training, and show in Sec.~\ref{Sec:Res} that the method can generalize to $W^{+}$ space to handle real images projected into StyleGAN2. In order to robustly train the network and avoid degenerate solutions (e.g., \alphanet to produce all 1s), we make several changes as described next. 

\textbf{(a) Discriminator \disc}: Unlike  using a frozen pre-trained generator ($\gen, \genBG$), we train the discriminator from scratch. Our method fails to converge by starting the training from a pre-trained discriminator. We hypothesize that that the pre-trained discriminator is already very strong in detecting the correlations between the foreground and the background~\cite{salimans2016improved}. For example, environmental illumination, shadows and reflections in case of cars, cats, and horses. Hence to enable our network to train on independently sampled backgrounds, we use a `Weak' discriminator, in the sense that it does not have a strong prior about the correlation between foregrounds and backgrounds.

\textbf{(b) Truncation trick}: We observed that unlike FFHQ trained StyleGAN2, the samples from the LSUN-Object trained StyleGAN2 have variable quality produced without truncation. To avoid the rejection sampling and ensure high quality samples during the training, we use truncation $\psi$ $\in$ [0,1] for the generator $\gen$ in case of LSUN-Object training~\cite{STYLEGAN2018}.

\textbf{(c) Regularizer}: As the alpha segmentation network may choose to converge into a sparse map, we use a binary enforcing regularizer, $B(M) := \min(M, 1-M)$ to guide the training. Note that we control this regularizer to still get soft segmentation maps. The truncation trick above may result in backgrounds from \genBG more aligned to the original distribution than the composite image from \gen. Hence, for the LSUN-Object, we use another regularizer $C(M) := \text{ReLU}(\phi_1 - \frac{1}{m\times m} \sum_{i} M_i )$ which ensures that the optimization does not converge to a trivial solution of only using the background network. Similarly, the regularizer as $E(M) := \text{ReLU}(\phi_2 - \frac{1}{m\times m} \sum_{i} (1- M_i) )$  checks that the solution does not degenerate to only using the \gen network.

\begin{table*}[thpb]
\caption{Evaluation of unsupervised $PSeg$~\cite{githubrepo2, bielski2019emergence} against our unsupervised approach using results from supervised network BiSeNet~\cite{Yu_2018_ECCV, githubrepo}, trained on CelebA-Mask dataset~\cite{CelebAMask-HQ}, as ground truth.  Note that for $PSeg$, since the samples can be of low-quality, we use the Detectron2 model for person detection before evaluating the masks. See Figure~\protect\ref{fig:fig1} for assessing the visual quality of the BiSeNet generated masks. } 
\begin{center}

\footnotesize
\begin{tabular}{rrrrrrrrrrrrrr} \toprule

\multirow{2}{*}{Method}   & \multicolumn{6}{c}{Truncation $\Psi$ = 0.7 } & \multicolumn{6}{c}{Truncation $\Psi$ = 1.0}\\  \cmidrule{2-14}
 \multicolumn{1}{c}{} &
 \multicolumn{1}{c}{IOU fg/bg}  & \multicolumn{1}{c}{mIOU}  & \multicolumn{1}{c}{F1} & \multicolumn{1}{c}{Prec} & \multicolumn{1}{c}{Rec} & \multicolumn{1}{c}{Acc} & 
 \multicolumn{1}{c}{IOU fg/bg}  & \multicolumn{1}{c}{mIOU}  & \multicolumn{1}{c}{F1} & \multicolumn{1}{c}{Prec} & \multicolumn{1}{c}{Rec} & \multicolumn{1}{c}{Acc} &
\hspace{0.15cm}
 \\ \midrule

$PSeg$ &0.52/0.82	&0.67	&0.80	&0.78	&0.81	  &0.85    &0.50/0.81   &0.66	&0.78	&0.77	&0.80	&0.84	     \\
$Ours$  &\textbf{0.87/0.94}	&\textbf{0.90}	&\textbf{0.95}	&\textbf{0.95}	&\textbf{0.94}	    &\textbf{0.95}
 &\textbf{0.75/0.89}	&\textbf{0.82}	&\textbf{0.90}	&\textbf{0.92}	&\textbf{0.89}	    &\textbf{0.92}\\

\bottomrule
\end{tabular}
\end{center}

\label{tab: table1}
\end{table*}

\begin{figure}
    \centering
    \includegraphics[width=\linewidth]{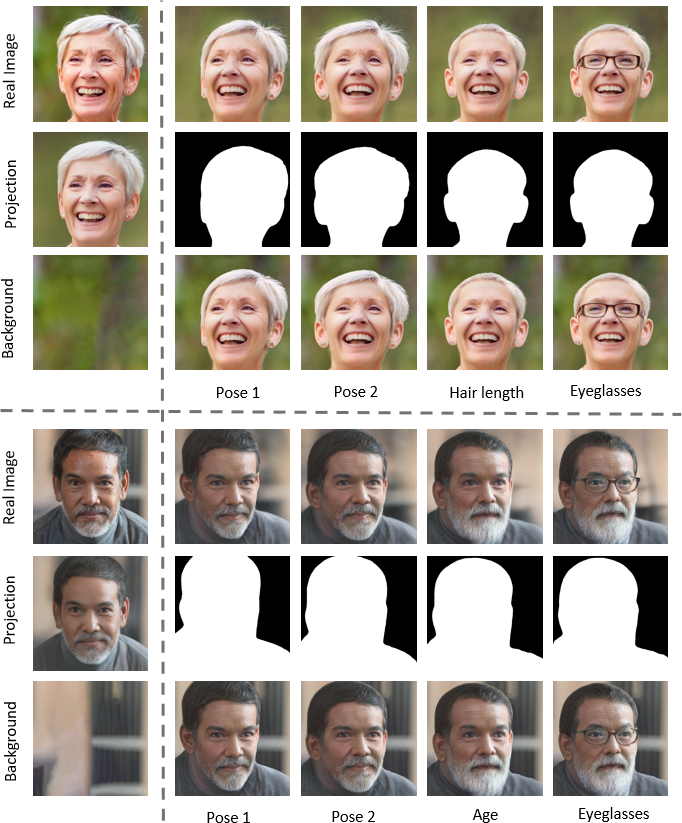}
    \caption{Our method achieves better background preservation compared to original semantic edits in StyleFlow~\cite{abdal2020styleflow}. For the real image, we first obtain a background layer, segmented using Label4Free and then completed using ContentAwareFill, and then for each edit using StyleFlow, we again segment them using our method and then composite back with the completed background layer (obtained above). Please compare the first row versus third rows.
    }
    \label{fig:fig4}
\end{figure}

\section{Results}
\label{Sec:Res}
\subsection{Evaluation Measures}
For the task of segmentation, IOU (Intersection Over Union) and mIOU (mean Intersection Over Union) give the measure of overlap between the ground truth and predicted segments. Since there are two classes we calculate IOU for both the classes in the experiments. We also report the final mIOU. Additional metrics that we report to evaluate the segmentation are Precision (Prec), Recall (Rec), F1 Score, and accuracy (Acc). As surrogate for visual quality, we use FID in our experiments. This can only be seen as a rough approximation to visual quality and is mainly useful in conjunction with visual inspection of the results.

\subsection{Datasets}

We use StyleGAN2 pretrained on FFHQ~\cite{STYLEGAN2018}, LSUN-Cars, LSUN-Cats, and LSUN-Horse~\cite{yu15lsun} datasets. FFHQ is a face image dataset with resolution $1024\times1024$. The facial features are aligned canonically. The LSUN-Object dataset contains various objects. It is relatively diverse in terms of poses, position, and number of objects in a scene. We use the subcategories for cars, cats, and horses.

\subsection{Competing Methods}
\label{Sec:comp}
We compare our results with two approaches, a supervised approach and an unsupervised approach. 

In the \textit{supervised setting}, we use BiSeNet~\cite{Yu_2018_ECCV, githubrepo} trained on the CelebA-Mask dataset~\cite{CelebAMask-HQ} for the evaluation of the faces and Facebook's Detectron 2 Mask R-CNN Model ( R101-FPN~\cite{githubrepo3} ) with the ResNet101 architecture pre-trained
on the MS-COCO dataset~\cite{lin2014microsoft} for the evaluation on LSUN-Object datasets. As our method is unsupervised, these methods are mainly suitable to judge how close we can come to supervised training. They are not direct competitors to our method. We also create a custom evaluation dataset of 10 images per class to directly compare the two approaches (See Supplemental (Appendix C)). 

In the \textit{unsupervised setting}, we compare our method with PSeg~\cite{githubrepo2, bielski2019emergence} using the parameters in the open source GitHub repo. This method is the only published competitor to ours.

\subsection{Comparison on Faces} 
\begin{figure}[t!]
    \centering
    \includegraphics[width=\linewidth]{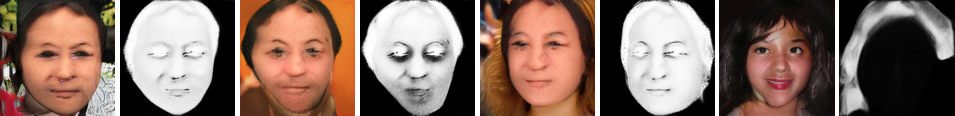}
    \caption{Samples from $PSeg$ method trained on FFHQ dataset. For samples on LSUN-Object refer to Figure 3 and Figure 4 in $PSeg$~\cite{bielski2019emergence} paper. Note that the method struggles with the quality of samples and the authors communicated that they are working on a better version.} 
    \label{fig:fig_pert}
\end{figure}

We compare and evaluate segmentation results on both sampled and real images quantitatively and qualitatively. First we show qualitative results of the learned segmentation masks using the images and backgrounds sampled from the StyleGAN2 in Fig.~\ref{fig:fig1}. To put these results into context, we compare the segmentation masks of the learned alpha network \alphanet with the \emph{supervised} BiSeNet.
The figure shows results at different truncation levels. The truncation setting $\psi=0.5$ leads to higher quality images. The setting $\psi=1.0$ leads to lower quality images that are more difficult to segment from the composited representation. In such cases our method is able to outperform the supervised segmentation network.
In Table~\ref{tab: table1},  we compare the results of the unsupervised segmentation of StyleGAN2 generated images using our method with unsupervised Pseg. We train PSeg~\cite{githubrepo2, bielski2019emergence} using the parameters in the Github repo on the FFHQ dataset. In the absence of a large corpus of testing data, we estimate the ground truth using a supervised segmentation network (BiSeNet). We sample $1k$ images and backgrounds at different truncation levels and compute the evaluation metrics. The quantitative results clearly show that our method is able to produce results of much higher quality than our only direct (unsupervised) competitor. The PSeg method tends to extract some random attributes from the learned images and is very sensitive to hyperparameters (See Fig.~\ref{fig:fig_pert}). In contrast, our method ensures that the original generative quality of StyleGAN is maintained. We also compare the FID of the sampled images in Table~\ref{tab: table3}. The results show that PSeg drastically affects the sample quality.

\paragraph{Background preserving face editing}
In order to show the application for our method in real image editing, we use state-of-the-art image editing framework StyleFlow~\cite{abdal2020styleflow}. One of the limitations of sequential editing using StyleGAN2 is that it has a tendency to change the background. Fig.~\ref{fig:fig4} shows some edits on the real and sample images using StyleFlow and corresponding results where our method is able to preserve the background using the same edit. Notice that our method is robust to pose changes. 

\paragraph{Performance on real face images.}
One method to obtain segmentation results on real images is to project them into the StyleGAN2 latent space. We do this for $1k$ images from the CelebA-Mask dataset using the PSP encoder~\cite{richardson2020encoding}. We show quantitative results in Table~\ref{tab: table2} comparing the output segmentation masks with the ground truth in the dataset. Note that the Alpha Network \alphanet is not trained on the $W^{+}$ space but is able to generalize well on the real images. A downside of this approach is that the projection method introduces its own projection errors affecting segmentation.

\paragraph{Labels4Free to obtain synthetic training data.}
A better method to extend to arbitrary images is to generate synthetic training data (see Table~\ref{tab: table2}). We train a UNet~\cite{ronneberger2015u} and BiSeNet on $10k$ sampled images and backgrounds produced by our unsupervised framework and report the scores. Note that the scores are comparable to the supervised setup. This demonstrates that our method can be applied to synthetically generate high quality training data for class specific foreground/background segmentation.

\begin{table}[t!]
\caption{Evaluation of unsupervised $PSeg$~\cite{githubrepo2, bielski2019emergence} against our unsupervised approach on LSUN-Object categories (LSUN-Horse, LSUN-Cat and LSUN-Car) using results from supervised network Detectron 2 R101-FPN~\cite{githubrepo3} trained on MS-COCO dataset. We report the results without truncation and the threshold for the masks is 0.9. A: LSUN-Cat ; B : LSUN-Horse ; C: LSUN-Car. } 
\centering
\scriptsize
\begin{tabular}{rrrrrrrr} \toprule

\multicolumn{1}{c}{Method}  & \multicolumn{1}{c}{IOU fg/bg}  & \multicolumn{1}{c}{mIOU}  &  \multicolumn{1}{c}{F1}& \multicolumn{1}{c}{Prec}& \multicolumn{1}{c}{Rec}& \multicolumn{1}{c}{Acc}\\ \midrule
$PSeg$(A)  &0.65/0.68	&0.66	&0.80	&0.81 &0.80 &0.79 \\
Ours(A) 
&\textbf{0.84/0.77}	&\textbf{0.81}	&\textbf{0.89}	&\textbf{0.89} &\textbf{0.90} &\textbf{0.90}
\\\midrule
$PSeg$(B)  &0.50/0.40	&0.45	&0.71	&0.69 &0.73 &0.63 \\
Ours(B) 
&\textbf{0.83/0.67}	&\textbf{0.75}	&\textbf{0.85}	&\textbf{0.84} &\textbf{0.91} &\textbf{0.87}
\\\midrule
$PSeg$(C)  &0.81/0.73	&0.77	&0.83	&0.83 &0.84 &0.85 \\
Ours(C)
&\textbf{0.93/0.84}	&\textbf{0.89}	&\textbf{0.94}	&\textbf{0.93} &\textbf{0.95} &\textbf{0.95}
\\
\bottomrule
\end{tabular}
\label{tab: table_others_sampled}
\end{table}

\begin{table}[thpb]
\caption{Evaluation on custom dataset. Dt2 : Detectron 2 A: FFHQ ; B: LSUN-Cat ; C: LSUN-Horse ; D: LSUN-Car.  } 
\centering
\scriptsize
\begin{tabular}{rrrrrrrr} \toprule

\multicolumn{1}{c}{Method}  & \multicolumn{1}{c}{IOU fg/bg}  & \multicolumn{1}{c}{mIOU}  &  \multicolumn{1}{c}{F1}& \multicolumn{1}{c}{Prec}& \multicolumn{1}{c}{Rec}& \multicolumn{1}{c}{Acc}\\ \midrule
BiSeNet(A) &0.94/0.97	&0.95	&\textbf{0.98}	&0.98 &0.97 &\textbf{0.98} \\
Ours(A)  &\textbf{0.95/0.98}	&\textbf{0.96}	&\textbf{0.98}	&\textbf{0.99} &\textbf{0.98} &\textbf{0.98}

\\\midrule
Dt2(B)   &0.93/0.95	&0.94	&0.97	&0.97 &0.97 &0.97 \\
Ours(B) 
&\textbf{0.95/0.96}	&\textbf{0.96}	&\textbf{0.98}	&\textbf{0.98} &\textbf{0.98} &\textbf{0.98}
\\\midrule
Dt2(C)  &0.96/0.94	&0.95	&0.97	&0.97 &0.97 &0.98 \\
Ours(C)  &\textbf{0.97/0.96}	&\textbf{0.97}	&\textbf{0.98}	&\textbf{0.98} &\textbf{0.98} &\textbf{0.98}
\\\midrule
Dt2(D)  &\textbf{0.99/0.96}	&\textbf{0.97}	&\textbf{0.99}	&\textbf{0.99} &0.98 &\textbf{0.99} \\
Ours(D)
&0.98/0.95	&\textbf{0.97}	&0.98	&0.98 &\textbf{0.99} &\textbf{0.99}
\\
\bottomrule
\end{tabular}
\label{tab: table_others}
\end{table}

\begin{table}[thpb]
\caption{Evaluation of the unsupervised object segmentation of real images using a projection method (PSP) and segmentation networks (Unet and BiSeNet) trained on our generated synthetic data. } 
\label{tab: table2}
\centering
\scriptsize
\begin{tabular}{rrrrrrrr} \toprule
 \multicolumn{1}{c}{Method}  &
 \multicolumn{1}{c}{IOU fg/bg}  & \multicolumn{1}{c}{mIOU}  &  \multicolumn{1}{c}{F1} & \multicolumn{1}{c}{Prec} & \multicolumn{1}{c}{Rec} & \multicolumn{1}{c}{Acc} \\\midrule
                       
BiSeNet & \textbf{0.84/0.92} &\textbf{0.88} &\textbf{0.93} &0.93 &\textbf{0.94} &\textbf{0.94} \\
Ours(PSP) &0.83/\textbf{0.92}	&\textbf{0.88}	&\textbf{0.93}	&\textbf{0.94} &0.93 &\textbf{0.94}
\\
Ours(UNet) &0.75/0.88	&0.82	&0.90	&0.90 &0.90 &0.91 \\
Ours(BiSeNet) &0.81/0.91	&0.86	&0.92	&0.92 &0.92 &0.93 \\
\bottomrule
% $Dt2$(B) &  && && & \\
% Ours(B-UNet) &0.83/0.59	&0.71	&0.83	&0.81 &0.84 &0.87
\\

\end{tabular}
\end{table}

\begin{table}[thpb]
\caption{FID comparison of the generation quality of samples from $PSeg$ method vs StyleGAN2. A : FFHQ ; B : LSUN-Cat ; C : LSUN-Horse ; D : LSUN-Car. } 
\centering
\scriptsize
\begin{tabular}{rrrrrr} \toprule
 \multicolumn{1}{c}{Method}  &
 \multicolumn{1}{c}{A}  & \multicolumn{1}{c}{B}  &  \multicolumn{1}{c}{C} & \multicolumn{1}{c}{D} &  \\\midrule

$PSeg$ &24.33 	&57.32	&53.60	&30.87 \\
StyleGAN2 &2.82	&6.93	&3.43	&2.32 
\\
\bottomrule
\end{tabular}
\label{tab: table3}
\end{table}

\subsection{Comparison on other datasets}

We also train our framework on LSUN-Cat, LSUN-Horse and LSUN-Car~\cite{yu15lsun}. These are more challenging datasets than FFHQ. We have identified two related problems with StyleGAN2 trained on these datasets. Firstly, the quality of the samples at lower truncation or no truncation levels is not as high as the FFHQ trained StyleGAN2. Second, these datasets can have multiple instances of the objects in a scene. StyleGAN2 does not handle such samples well. Both these factors affect the training as well as the evaluation of the unsupervised framework.

In Fig.~\ref{fig:fig_lsun_object} we show the qualitative results of our unsupervised segmentation approach on LSUN-Object datasets. Notice that the quality of segmentation masks produced by our framework is comparable to the supervised network. In Table~\ref{tab: table_others_sampled}, we calculated the quantitative results of $1k$ sampled images from StyleGAN2. Note that here we resort to rejection sampling based on the technique to reject the sample not identified by the detectron 2 model. For simplicity we also reject the multiple instance object samples. The results show that the metrics scores are comparable to a state-of-the-art supervised method. 
In order to directly compare the supervised and unsupervised approaches we compare the metrics on a custom dataset mentioned in Section~\ref{Sec:comp}. Table~\ref{tab: table_others} shows that our supervised approach is either better or has similar segmentation capabilities as the supervised approach.

%-------------------------------------------------------------------------
\subsection{Training Details}
Our method is faster to train than the competing methods. We train our framework on 4 RTX 2080 (24 GB) cards with a batch size of 8 using the StyleGAN pytorch implementation~\cite{strepo}. Let $\lambda_1$ and $\lambda_2$ be the weights of the regualarizers $B(M)$ and $C(M)$,  respectively. For FFHQ dataset, we run $1k$ iterations and set $\lambda_1 = 1.2$. For LSUN-Object, we set $\phi = 0.25$. For LSUN-Cat, we run 900 iterations and set  $\lambda_1 = \lambda_2 = 3$ and  $\psi = 0.5$.  For LSUN-Horse, we run 500 iterations and set  $\lambda_1 = \lambda_2 = 3$ and  $\psi = 1.0$. For LSUN-Car, we run 250 iterations and set  $\lambda_1 = \lambda_2 = 20$ and  $\psi = 0.3$. Also, we set the weight of the non-saturating discriminator loss to $0.1$. For FFHQ, LSUN-Horse and LSUN-Cat, we set the learning rate to $0.0002$. For LSUN-Car, we set the learning rate to $0.002$.  Each model takes less than 30 minutes to converge.

\section{Conclusion}
We proposed a framework for unsupervised segmentation of StyleGAN generated images into a foreground and a background layer. The most important property of this segmentation is that it works entirely without supervision. To that effect, we leverage information that is already present in the layers to extract segmentation information.
In the future, we would like to explore the unsupervised extraction of other information, e.g., illumination information, the segmentation into additional classes, and depth information.
\section{Acknowledgements}
This work was supported by Adobe and the KAUST Office of Sponsored Research (OSR) under Award No. OSR-CRG2017-3426.

{\small
\bibliographystyle{ieee_fullname}
\bibliography{egbib}
}

\section{Appendix A}
\subsection{Ablation Study }
In order to validate the importance of the in-domain backgrounds for the training of the unsupervised network, we train our framework with the backgrounds taken from the MIT places~\cite{NIPS2014_3fe94a00} dataset. In order to do so, we replace the background generator with a random selection of an image from MIT places. During training, we not only train the alpha network but also the discriminator (as in the main method described in the paper). As discussed in the main paper, the discriminator is very good in identifying the out of the distribution images. In Table~\ref{tab: table_mit}, we show the  scores compared with BiSeNet and Detectron 2 (see Table 1 and Table 2 in the main paper) when using the MIT places for the backgrounds. Note that the scores decrease drastically.

\begin{table}[h!]
\caption{ Quantitative results of using MIT places dataset for the backgrounds.  } 
\centering
\small
\begin{tabular}{rrrrrrr} \toprule
 \multicolumn{1}{c}{Dataset}  &
 \multicolumn{1}{c}{mIOU}  &  \multicolumn{1}{c}{F1} & \multicolumn{1}{c}{Prec} & \multicolumn{1}{c}{Rec} & \multicolumn{1}{c}{Acc} \\\midrule
                       
FFHQ &0.34 &0.40 &0.34 &0.50 &0.68  \\
LSUN-Horse &0.14	&0.22	&0.14	&0.5 &0.28
\\
LSUN-Cat &0.20	&0.28	&0.69	&0.50 &0.39 \\
LSUN-Car &0.34	&0.51	&0.63	&0.63 &0.51  \\
\bottomrule

\\

\end{tabular}
\label{tab: table_mit}
\end{table}

As a second ablation study, we try to learn the Alpha mask only from features of the last layer before the output. This straightforward extension does not work well as seen in Table~\ref{tab: table_supp_layer}. In summary, using features from multiple layers in the generator is important to achieve higher fidelity.

\begin{table}[h!]
\caption{ Quantitative results of using only the last layer of the StyleGAN2 for the construction of the Alpha Network.  } 
\centering
\small
\begin{tabular}{rrrrrrr} \toprule
 \multicolumn{1}{c}{Dataset}  &
 \multicolumn{1}{c}{mIOU}  &  \multicolumn{1}{c}{F1} & \multicolumn{1}{c}{Prec} & \multicolumn{1}{c}{Rec} & \multicolumn{1}{c}{Acc} \\\midrule
                       
FFHQ &0.34 &0.41 &0.57 &0.50 &0.69  \\
LSUN-Horse &0.33	&0.42	&0.41	&0.44 &0.63
\\
LSUN-Cat &0.31	&0.41	&0.48	&0.50 &0.58 \\
LSUN-Car &0.42	&0.57	&0.57	&0.58 &0.63  \\
\bottomrule

\\

\end{tabular}
\label{tab: table_supp_layer}
\end{table}

\section{Appendix B}
\subsection{Visualization of tRGB layers}
In Fig.~\ref{fig:viz}, ~\ref{fig:viz_horse} and ~\ref{fig:viz_car},  we visualize the tRGB layers at different resolutions as discussed in the experiment in Section 3.1 of the main paper. We select all the resolution tensors including and after the highlighted resolution. Here, we first normalize the tensors by $\frac{x - \min(x)}{\max(x)-\min(x)}$, where $x$ represents the tRGB tensor at a given resolution. Notice that for the face visualization in Fig.~\ref{fig:viz}, the face structure is clearly noticeable at the $32\times32$ resolution corresponding to the 4th layer of StyleGAN2. Other efforts in the StyleGAN-based local editing~\cite{abdal2020image2stylegan++, wu2020stylespace} also selects early layers for the semantic manipulation of the images. These tests support that even features from earlier layers are beneficial for segmentation.
\begin{figure}[h]
    \centering
    \includegraphics[width=0.7\columnwidth]{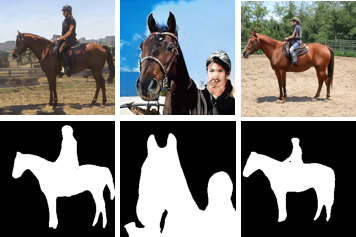}
    \caption{Multiple object segmentation.
    }
    \label{fig:multiple}
\end{figure}

\begin{figure*}[t!]
    \centering
    \includegraphics[width=\linewidth]{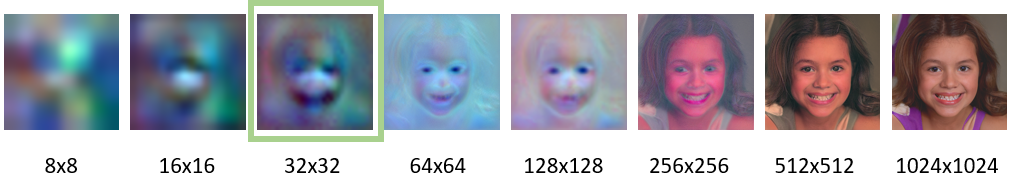}
    \caption{Visualization of the tRGB layers in the StyleGAN2 trained on FFHQ dataset. Note that the maps produce a prominent face structure at $32\times32$ resolution corresponding to layer 4 of StyleGAN2. 
    }
    \label{fig:viz}
\end{figure*}

\begin{figure*}[t!]
    \centering
    \includegraphics[width=\linewidth]{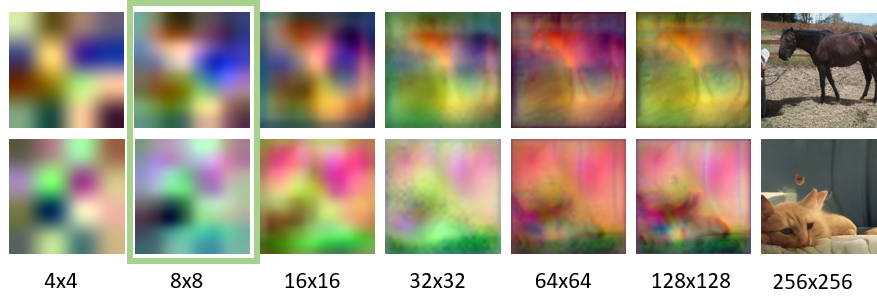}
    \caption{Visualization of the tRGB layers in the StyleGAN2 trained on LSUN-Horse and LSUN-Cat datasets.  
    }
    \label{fig:viz_horse}
\end{figure*}

\begin{figure*}[t!]
    \centering
    \includegraphics[width=\linewidth]{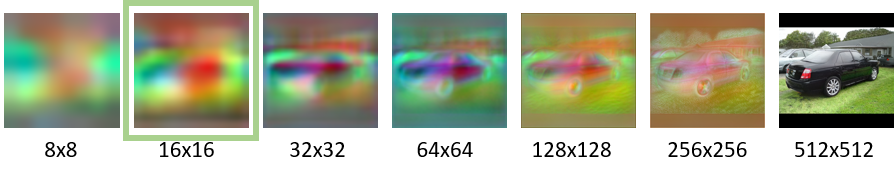}
    \caption{Visualization of the tRGB layers in the StyleGAN2 trained on LSUN-Car dataset.
    }
    \label{fig:viz_car}
\end{figure*}

\subsection{Multiple object segmentation}
Apart from the main object, there can be multiple instances of the same object or multiple objects in the scene in correlation with the foreground object. Fig.~\ref{fig:multiple} shows that our method is able to handle such cases.

\section{Appendix C}

\subsection{Custom dataset}
\begin{figure*}[t!]
    \centering
    \includegraphics[width=\linewidth]{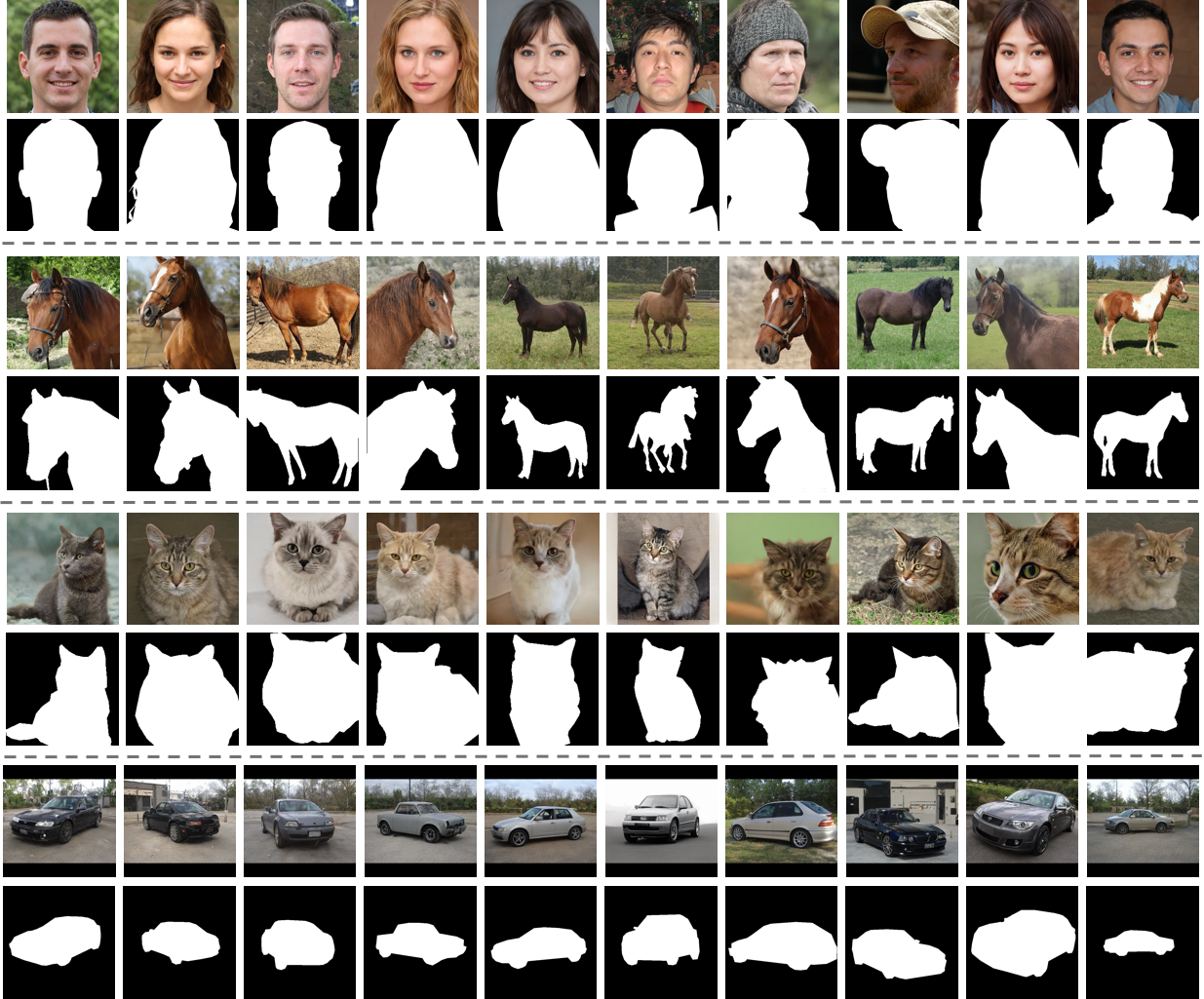}
    \caption{ Our custom dataset with the corresponding ground truth labels.
    }
    \label{fig:dataset}
\end{figure*}

We curated a custom dataset to evaluate the performance of our method with the supervised frameworks (BiSeNet and Detectron 2). We collected 10 images per class. These images were sampled by using the pretrained StyleGAN2 at different trucation levels and on different datasets i.e., FFHQ, LSUN-Horse, LSUN-Cat and LSUN-Car. Note that we collected a diverse set of images , \eg, diverse poses, lighting, and background instances (see Fig.~\ref{fig:dataset}). The images were annotatd using the LabelBox tool~\cite{LabelBox}.

\end{document}